\documentclass[10pt,twocolumn,letterpaper]{article}

\usepackage[pagenumbers]{wacv} 

\usepackage[dvipsnames]{xcolor}
\usepackage{amsmath}
\usepackage{pifont}
\usepackage{diagbox}

\usepackage{graphicx}
\usepackage{amsmath}
\usepackage{amssymb}
\usepackage{booktabs}

\DeclareMathOperator*{\argmax}{arg\,max}

\usepackage[pagebackref,breaklinks,colorlinks]{hyperref}

\usepackage[capitalize]{cleveref}
\crefname{section}{Sec.}{Secs.}
\Crefname{section}{Section}{Sections}
\Crefname{table}{Table}{Tables}
\crefname{table}{Tab.}{Tabs.}

\def\wacvPaperID{965} 
\def\confName{WACV}
\def\confYear{2025}

\begin{document}

\title{UnLearning from Experience to Avoid Spurious Correlations} 

\author{Jeff Mitchell\\
Queen's University Belfast\\
School of EEECS\\
United Kingdom \\
{\tt\small jmitchell25@qub.ac.uk}
\and
Jesús Martínez del Rincón\\
Queen's University Belfast\\
School of EEECS\\
United Kingdom \\
{\tt\small j.martinez-del-rincon@qub.ac.uk}
\and
Niall McLaughlin\\
Queen's University Belfast\\
School of EEECS\\
United Kingdom \\
{\tt\small n.mclaughlin@qub.ac.uk}
}
\maketitle

\begin{abstract}
    While deep neural networks can achieve state-of-the-art performance in many tasks, these models are more fragile than they appear. They are prone to learning spurious correlations in their training data, leading to surprising failure cases. In this paper, we propose a new approach that addresses the issue of spurious correlations: UnLearning from Experience (ULE). Our method is based on using two classification models trained in parallel: student and teacher models. Both models receive the same batches of training data. The student model is trained with no constraints and pursues the spurious correlations in the data. The teacher model is trained to solve the same classification problem while avoiding the mistakes of the student model. As training is done in parallel, the better the student model learns the spurious correlations, the more robust the teacher model becomes. The teacher model uses the gradient of the student's output with respect to its input to unlearn mistakes made by the student. We show that our method is effective on the Waterbirds, CelebA, Spawrious and UrbanCars datasets.
\end{abstract}

\section{Introduction}
\label{sec:introduction}


Training Deep Learning (DL) models is a well-studied problem that usually involves minimizing the average loss on the training set. The underlying assumption is that the data in the training and testing sets are drawn from identical distributions. However, in many realistic situations, the training set does not reflect the full diversity of realistic test data. Therefore, the trained system does not generalise well to Out-Of-Distribution (OOD) or group-shifted data. This can happen because the trained system relies on various spurious correlations present in the training set but not present in the testing data, leading to performance drops in realistic settings.

\begin{figure}[t!]
    \centering
    \includegraphics[trim={3cm 4cm 10cm 2cm},clip,width=\columnwidth]{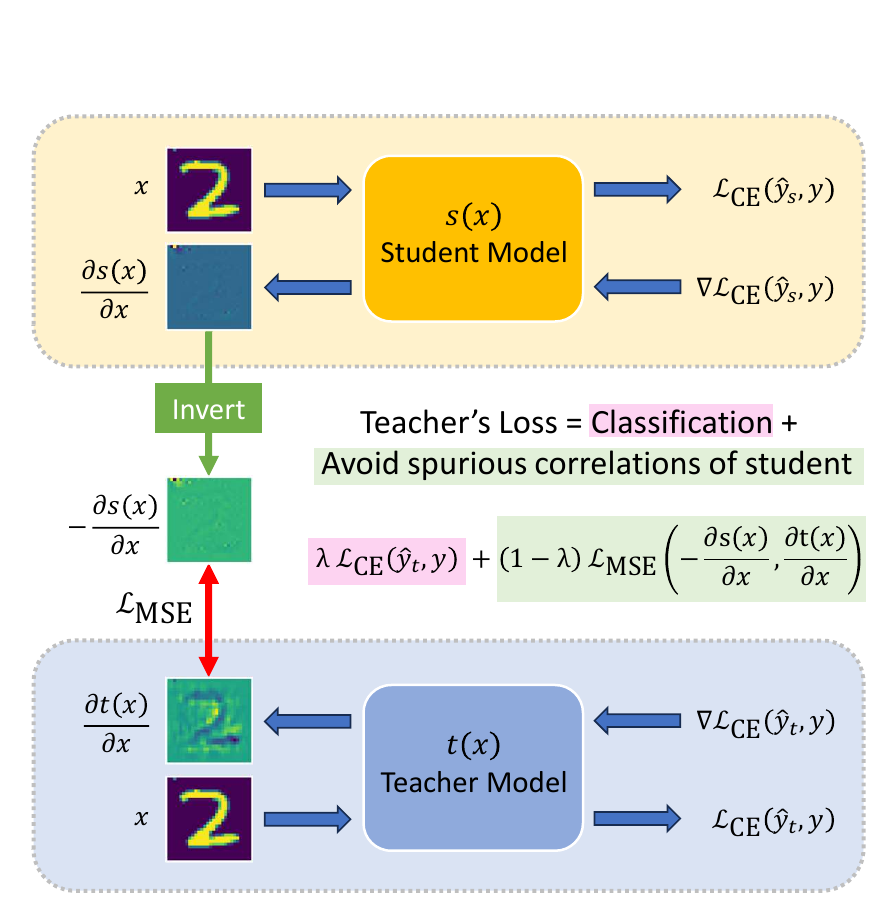}
    \caption{UnLearning from Experience (ULE). Overview of our proposed method. Two models are trained in parallel. The student model learns the spurious correlations, which the teacher model unlearns from the mistakes made by the student. \label{fig:system_flowchart}}
    \vspace{-0.5cm}
\end{figure}

Spurious correlations happen when, for a given dataset, there is a coincidental correlation between a non-predictive feature of the input and the label. If those spurious correlations are present during training, machine learning models may learn to use the non-predictive feature to solve the task. Then, when tested on the same task but without the spurious feature present, the system's performance will drop. For example, in the Waterbirds dataset~\cite{wah2011caltech,zhou2018places}, an image's background is correlated with the label. A model could learn to associate the presence of water backgrounds with the label water-bird and land backgrounds with the label land-bird rather than looking at the bird to solve the task. In the simplest case, a dataset may have one spurious correlation. In more complex scenarios, the term group shift is used to describe cases with multiple sub-groups within the dataset, each of which may be subject to multiple different spurious correlations.

Many existing methods for spurious correlation robustness explicitly use group labels during training~\cite{sagawa2019distributionally,nam2020learning,pmlr-v139-creager21a,pmlr-v139-liu21f,nam2021spread}.
In practice, full details of the number and kinds of spurious correlations and/or explicit group label information may not be available.
We propose, UnLearning from Experience (ULE), which allows the creation of a model robust to spurious correlations and does not require any group label information at either training or testing time. In our approach, two models are trained in parallel. A student model $s(x)$ is directly trained on the dataset. A teacher model $t(x)$ then unlearns spurious correlations by observing the gradient of the student's output $\partial s(x) / \partial x$ with respect to its input $x$. Thus, the teacher model learns to avoid the mistakes made by the student, while also solving the task of interest. \Cref{fig:system_flowchart} shows an overview of our proposed method. We demonstrate the validity of this approach for classification tasks.
The main contributions of this paper are:
\begin{itemize}
    \item We propose a new twist on student-teacher methods by reversing the traditional student and teacher roles to improve spurious correlation robustness.
    \item We train the two models in parallel, optimising a loss where the teacher looks at the student's gradient to avoid repeating the student's mistakes.
    \item We do not require knowledge of the presence of spurious correlation or group-shifts. Group labels are not required at training or testing time.
    \item We use XAI to show how our model avoids learning spurious correlations.
    \item Our method achieves SOTA results on Waterbirds and CelebA. And comparable results on Spawrious. 
    
    
\end{itemize}

\section{Related Work}
\label{sec:literature_review}

%

The standard neural network training approach is Empirical Risk Minimization (ERM)~\cite{NIPS1991_ff4d5fbb}, which minimizes the average loss on the training data. 
ERM does not confer robustness to spurious correlations or group shifts. Recently, various approaches have been proposed to increase robustness to these effects~\cite{srivastava2023mitigating,joshi2023towards}.




\textbf{Group Labels used in Training} Common approaches to improving group-shift robustness use group labels during training and validation.
Sagawa et al.~\cite{sagawa2019distributionally} have shown that using Distributionally Robust Optimization (Group-DRO) to minimize the worst-group loss, coupled with strong $L_2$ regularization, results in models with high average test accuracy and high Worst-Group Accuracy (WGA). WGA has become established as the reference for validating spurious correlation robustness. However, DRO requires full supervision with explicit group labels, which is undesirable. Izmailov et al.~\cite{izmailov2022feature} studies the quality of features learned by ERM. They find that many robustness methods work by learning a better final layer. A similar approach is taken by Kirichenko et al.~\cite{kirichenko2022last} where the last layer features are reweighted using a small dataset without the spurious correlation present. Qiu et al.~\cite{qiu2023afr} again re-trained the final layer using a weighted loss that emphasizes samples where ERM predicts poorly.

\textbf{Explicit Group Labels in Validation} 
More flexible approaches to improving group robustness do not require group labels during training. Instead, a small sample of group labels, usually explicitly provided, are present in the validation set. Nam et al.~\cite{nam2020learning} train debiased models from biased models via their Learning from Failure (LfF) framework. They intentionally train a biased model and amplify its prejudice. They then train a debiased model that learns from the mistakes of the biased model by optimizing a generalized cross-entropy loss to amplify the bias. Similarly, Zhang et al.~\cite{zhang2018generalized} improve robustness to OOD and noisy datasets using a noise-robust generalized cross-entropy loss.

Group labels and pseudo-labels can be inferred from the data as demonstrated by Environment Inference for Invariant Learning (EIIL)~\cite{pmlr-v139-creager21a}, a general framework for domain-invariant learning that directly discovers partitions from the training data that are maximally informative for downstream invariant learning. Spread Spurious Attribute (SSA)~\cite{nam2021spread} is a semi-supervised method that leverages samples with/without spurious attribute annotations to train a model that predicts the spurious attribute. It then uses pseudo-labelled examples to train a new robust model.
%
%
Xie et al.~\cite{xie2020self} propose NoisyStudent (NS), a semi-supervised self-learning method which first trains a supervised teacher model to generate pseudo-labels for an equal or larger student model trained on the data with noise injected. It makes use of data augmentation, dropout and stochastic depth. This method is iterated multiple times by making the student model the new teacher and repeating. In Just Train Twice (JTT)~\cite{pmlr-v139-liu21f}, a model is trained for a small number of epochs; then a second model is trained by up-weighting samples where the first model has made a mistake.
Lee et al.~\cite{lee2022diversify} train an ensemble of models with a shared backbone. The models are forced to be diverse during training, and a final robust model is selected by observing a small number of new samples. Pagliardini et al.~\cite{pagliardini2022agree} take a similar approach of training an ensemble of models to agree on the training data but give different predictions on OOD data.
Finally, we note that related student-teacher approaches have been proposed for machine-unlearning in a more general context~\cite{kim2022efficient}.

\textbf{No Explicit Group Labels} Group labels may be scarce in real-world settings. 
Invariant Risk Minimization (IRM)~\cite{arjovsky2019invariant} assumes that training data is collected from separate distinct environments. IRM promotes learning a representation with correlations that are stable across these environments. CORrelation ALignment (CORAL)~\cite{coral} for unsupervised domain adaptation aims to minimize the domain shift by aligning the second-order statistics of the source and target distributions without requiring any target labels. Deep-CORAL~\cite{dcoral} extends this technique to deep neural networks. CausIRL~\cite{chevalley2022invariant} takes a causal perspective on invariant representation learning by deriving a regularizer to enforce invariance through distribution matching. Adversarial feature learning can also help with tackling the problem of domain generalization~\cite{Li_2018_CVPR}, by using Adversarial Autoencoders (MMD-AAE)  trained with a Maximum Mean Discrepancy (MMD) regularizer to align the distributions across different domains.

Finally, approaches have been proposed for producing robust models without group labels or the need for explicit domain generalization. Mehta et al.~\cite{mehta2022you} extract embeddings from a large pre-trained Vision Transformer, then train a linear classification layer using these embeddings. This approach does not require group labels, although it primarily relies on the pre-existing robustness of the embeddings from the pre-trained model~\cite{darbinyan2023identifying}. Tiwari et al.~\cite{pmlr-v202-tiwari23a} use an auxiliary network to identify and erase predictive features from lower network layers. Zhao et al.~\cite{NEURIPS2022_baaa7b5b} suppress shortcuts during training using an autoencoder, thus improving generalization. Zhang et al.~\cite{zhang2022correct} trains an ERM model to identify samples of the same class but with different spurious features, then uses contrastive learning to improve representation alignment. Recently Yang et al.~\cite{yang2024identifying} proposed SePArate early and REsample (SPARE), which identifies spuriously correlated samples early in training and uses importance sampling to minimize their effect. It does not require a group-labelled validation set. 

In common with many of the above approaches, ULE only uses group labels in the validation set to select network hyperparameters, which is also done by all rival methods (JTT, SSA, LfF, EIIL, Group DRO). Additionally, SSA, JTT and Groups DRO require group labels during training, for their methods to work. ULE and other methods (EIIL, LfF) only need group labels in validation for tuning hyperparameters. The methods JTT~\cite{pmlr-v139-liu21f}, LfF~\cite{nam2020learning} and NS~\cite{xie2020self} are the most similar to our proposed method as they use a second model to gain further insight into the dataset. However, these methods rely on generating pseudo-labels during validation. In contrast, our method does not use explicit group labels. In our method, one model observes the gradients of the other model to counteract spurious correlations. The key advantage of ULE over our closest performing rivals, SSA and JTT, is that once the hyperparameters of ULE are set, ULE does not require explicit group labels to compensate for spurious correlations. This makes ULE more general and elegant than SSA and JTT, which require more domain knowledge, i.e., labelled examples, to work.

\section{UnLearning from Experience}
\label{sec:method}

In this section, we introduce our proposed approach, UnLearning from Experience (ULE). As illustrated in~\Cref{fig:system_flowchart}, we train two models in parallel: a student model and a teacher model. The student model is trained to solve a classification task as normal, while the teacher model is trained in parallel, using a custom loss function, to solve the same classification task while avoiding spurious correlations learned by the student model. Both models are trained simultaneously, with identical batches, and their parameters are updated in parallel.

The core idea is that the student model will be prone to use shortcuts or spurious correlations in the dataset to solve the classification task. The teacher model is then trained to solve the same classification task with an additional term in its loss function to encourage it to avoid learning the same behaviour as the student model, hence avoiding the shortcuts or spurious correlations learned by the student model.

We purposefully reverse the names in our student-teacher paradigm. We want to emphasise the fact that the teacher is learning \textit{not} to copy the student, i.e., it is unlearning from the experience of the student.

It has been shown to be theoretically impossible to mitigate shortcuts without prior assumptions about the data~\cite{lin2022zin}. In practice, such mitigation is only possible when an assumption, such as simplicity bias, is imposed~\cite{li2023whac, yang2024identifying, shah2020pitfalls}. Thus, our underlying assumption is that learning short-cuts or spurious correlations is easier than learning the primary task. Therefore, it is more difficult to unlearn the correct semantic features.


Assume a classification function $s(x)$ that maps from an input image $x$ to the $\argmax$ class, $c$, of a normalised probability distribution, $\hat{y}_s$, over the classes. We will refer to $s(x)$ as the student network, trained using a conventional classification loss, such a cross-entropy $\mathcal{L}_{\text{CE}}(\hat{y}_s,y)$, where $y$ is the target probability distribution over the classes. We can define, $g_s$, a \emph{saliency map} for $s(x)$ as:
\begin{align}
    \label{eq:simple_gradients}
    g_s = \frac{\partial s(x)}{\partial x}
\end{align}
In other words, the saliency map is defined as the gradient of $s(x)$ with respect to the input $x$. It indicates parts of the input that, if changed, would affect the classification decision. 
We expect that any spurious correlations present will play an important part in the network's decision-making process. Hence, they will be highlighted in $g_s$. 
During training, averaged over all batch updates, the saliency map is expected to be dominated by meaningful features. Noise and features with very small magnitudes, which do not strongly influence the classification decision, will average out.
Therefore, $g_s$ can guide the training of another network to avoid following the same spurious correlations. 

In parallel with the student classification network, we train a teacher classification network, $t(x)$, using a loss function that includes both a standard classification loss and an additional term that causes it to avoid the spurious correlations highlighted in $g_s$. 
%
%
To encourage the teacher network to avoid spurious correlations, we use the loss term $\mathcal{L}_{\text{MSE}}$ to encourage the saliency map of the teacher network to be the opposite of the saliency map from the student network. $\mathcal{L}_{\text{MSE}}$ is defined as:
%
%
\begin{align}
    \label{eq:mse_loss}
    \mathcal{L}_{\text{MSE}}(-g_s,g_t) = \Vert \mathcal{N}(-g_s) - \mathcal{N}(g_t) \Vert_2^2
\end{align}
where 
$g_t = \partial t(x) / \partial x$ is the \emph{saliency map} of the teacher network. The function $\mathcal{N}(z)$ normalises the saliency maps to have a maximum value of $1$ by dividing the input by its maximum absolute value i.e.,
$\mathcal{N}(z) = z / \max(\lvert z \rvert)$. 

Note the term $-g_s$ in~\Cref{eq:mse_loss}. By multiplying $g_s$ by $-1$, the saliency map from the student network is inverted. Recall that our overall goal is to use $g_s$ to guide the training of the teacher network to avoid spurious correlations. In other words, features assigned high importance by the student network may coincide with spurious correlations; hence, the teacher network should try to assign low importance to these areas and vice versa. Thus, the $\mathcal{L}_{\text{MSE}}$ term encourages the saliency maps of both networks to be opposites. The final loss value is calculated via the mean squared difference between the two vectors. Pseudocode for UnLearning from Experience (ULE) is included in the supplementary material. 
The overall loss function for the teacher network is defined as follows:
%
\begin{align}
    \label{eq:overall_loss_function}
    \mathcal{L}_{\text{total}} = \lambda \mathcal{L}_{\text{CE}}(\hat{y}_t,y) + (1-\lambda) \mathcal{L}_{\text{MSE}}(-g_s,g_t) 
\end{align}
where $\mathcal{L}_{\text{CE}}$ is a classification loss, such a cross-entropy. The loss terms, $\mathcal{L}_{\text{CE}}$ and $\mathcal{L}_{\text{MSE}}$, are normalised to the same order of magnitude, then the hyperparameter $\lambda \in [0,1]$ balances the two loss terms. 







\subsection{Practical Implementation}
\label{sec:efficient_computation}

In the section above, we assume that saliency is based on $\partial t(x) / \partial x$, which is taken with respect to the input. Instead, we can freeze early layers of the network, treating them as a feature extractor, and compute $\partial t(x) / \partial A_j$, where $A_j$ is the matrix of activations at an intermediate layer $j$. 
%
%
It has been shown that modifying only the final layer of a pre-trained network can increase robustness to spurious correlations~\cite{kirichenko2022last,qiu2023afr,izmailov2022feature}. Moreover, we argue that features originally intertwined at the input may be separable at the final layer(s) of a pre-trained network, helping our approach to cope with more complex spurious correlations~\cite{sharif2014cnn}.

If the student and teacher networks have different architectures, there may be no layers with equal dimensionality. Therefore, a different way to perform supervision is needed. 
Let $A_j \in \mathbb{R}^{b\times d_j}$ be the activation matrix for a given hidden layer of the teacher network, where $b$ is the batch dimension, and $d_j$ is the flattened layer dimension. We can form the matrix $E_t=A_jA_j^{\mathsf{T}}$, where  $E_t \in \mathbb{R}^{b\times b}$, i.e., the size of $E_t$ is independent of the hidden layer dimensionality. The same process can be applied to any hidden layer, $A_k$ of the student network, to form matrix $E_s \in \mathbb{R}^{b\times b}$. In our training scheme, both networks are always trained in parallel with the same batch, so matrices $E_t$ and $E_s$ will always have the same size. The matrices $E_t$ and $E_s$ can be flattened and compared using $\mathcal{L}_{\text{MSE}}(-\mathcal{F}(E_s),\mathcal{F}(E_t))$ (see~\Cref{eq:mse_loss}), where $\mathcal{F}$ is the flattening operation, thus encouraging the student and teacher hidden layers to diverge. 

In practice (See Section~\ref{subsec:mixed_models}), we select $A_j$ and $A_k$ as the final fully connected layers of the teacher and student networks. Earlier layers are frozen. We note that last layer re-training is common in the spurious correlation literature~\cite{joshi2023towards,kirichenko2022last}. Our method then simplifies to training the final linear layer using $\mathcal{L}_{\text{total}}$ (see~\Cref{eq:overall_loss_function}).

\section{Experiments}
\label{sec:experiments}

In this section, we experimentally evaluate our proposed method qualitatively and quantitatively. We test our approach using several pre-trained models, including ResNet-18~\cite{he2016deep} pre-trained on the ImageNet1K\_V1, ResNet-50~\cite{he2016deep} pre-trained on ImageNet1K\_V2, and ViT-H-14~\cite{dosovitskiy2020image} comprising the original frozen SWAG~\cite{singh2022revisiting} trunk weights with a final linear classifier trained on  ImageNet1K.
As mentioned in \Cref{sec:efficient_computation}, we only fine-tune the final linear layer of each model. Unless otherwise noted, we use the same model for the student and teacher in all experiments. In all experiments, we train the models for $300$ epochs.

%
We tune our hyperparameters separately for each dataset and model by grid search. We vary the value of $\lambda$ in steps of 0.1 over the range [0, 1], the learning rate in the range [1e-1, 1e-5] in powers of 10, and select between standard and strong $L_2$ regularization. We select the values of $\lambda$, learning rate, and regularization that achieve the highest worst-group accuracy (WGA) on the validation set. We evaluate the model on the validation set every 10 epochs to monitor the model. We investigate the effect of strong $L_2$ regularization, which can be used to reduce the effects of spurious correlations by preventing the model from overfitting~\cite{sagawa2019distributionally}.

\subsection{Datasets}
\label{subsec:datasets}The following datasets, containing group-shifted and OOD data, were used to evaluate our proposed method: Waterbirds~\cite{wah2011caltech,zhou2018places}, CelebA~\cite{liu2015faceattributes} and Spawrious~\cite{lynch2023spawrious}. 


\textbf{Waterbirds Dataset}~\cite{sagawa2019distributionally}
 Consists of images of birds (land and water) cropped from the Caltech-UCSD Birds-200-2011 (CUB-200-2011) dataset~\cite{wah2011caltech} imposed onto backgrounds from the Places dataset~\cite{zhou2018places}. The resulting dataset contains $\approx$\,11,800 custom images. The objective is to classify the images into two classes: $y=\:$\{Waterbirds, Landbirds\} given spurious correlation, $a=\:$\{Water, Land\}, between the background and the bird class.

The test set consists of images from all four combinations of labels $y$ and spurious correlations $a$. To prevent bias in evaluating the model, we ensure that the number of test images in each group is balanced.
We calculate two main metrics to evaluate the robustness of models on the group-shifted data. The first metric is the average test accuracy, which is the average accuracy over all groups.
%

\begin{equation}
    \label{eq:average_test_accuracy}
    \overline{\text{Acc}} = \frac{1}{N_g} \sum_{i=1}^{N_g} \text{Acc}_i
\end{equation}
where $N_g=|y|*|a|$ is the number of groups and $\text{Acc}_i$  
is the model's accuracy on group $i$, calculated as the number of correct predictions divided by the total number of predictions for that group. However, the average test accuracy does not consider the distribution of the groups. For example, suppose the model performs well on the majority groups but poorly on the minority groups. In that case, the average test accuracy may be high, but the model will not be robust to group-shifted data. To measure robustness against spurious correlation, we look at WGA, defined as the model's accuracy on the worst-performing group.
%
\begin{equation}
    \label{eq:worst_group_accuracy}
    \text{WGA} = \min \{\text{Acc}_1, \text{Acc}_2, \dots, \text{Acc}_{N_g}\}
\end{equation}

WGA is the main metric in the literature \cite{sagawa2019distributionally,lynch2023spawrious,chevalley2022invariant,Li_2018_CVPR} rather than $\overline{\text{Acc}}$, as it specifically demonstrates robustness to spurious correlations.

%
%
\textbf{CelebA Dataset}
\label{subsec:celeba_dataset}
%
%
Following~\cite{sagawa2019distributionally}, we train models to classify images into two classes: $y=\:$\{Blond hair, Non-blond hair\} with a spurious correlation $a=\:$\{Female, Male\} between gender presentation and hair colour. We evaluate using WGA. The dataset has $\approx$\,202,600 images and balanced testing groups.

\textbf{UrbanCars Dataset}
\label{subsec:urbancara_dataset}
UrbanCars~\cite{li2023whac} include multiple types of spurious correlations. The task is to classify images into two classes: $y=\:$\{urban, country\}, given two types of spurious correlations, the background and a co-occurring object, which are correlated with the true class, and which both also take the classes $a=\:$\{urban, country\}.




\label{subsec:spawrious_dataset}
\textbf{Spawrious Dataset}~\cite{lynch2023spawrious} contains six datasets with easy, medium and hard variants of one-to-one and many-to-many spurious correlations. Many-to-many spurious correlations are more complex than the one-to-one spurious correlations in Waterbirds or CelebA.
Each dataset shows various dog breeds on different backgrounds. 
We classify images into four classes: $y=\:$\{Bulldog, Corgi, Dachshund, Labrador\} with a spurious correlation $a=\:$\{Desert, Jungle, Dirt, Mountain, Snow, Beach\} between the background and dog breed. Testing sets are balanced. Following the procedure in~\cite{lynch2023spawrious}, we evaluate using average accuracy and additionally with WGA for consistency with Waterbirds and CelebA.

\subsection{Proof of Concept}
\label{subsec:proof_of_concept}
We first demonstrate the effectiveness of our proposed ULE method using two modified versions of the MNIST dataset with artificial spurious correlations~\cite{deng2012mnist}. 
\textbf{MNIST-SC} - MNIST modified by one-hot encoding the class label into the upper left corner of every image. 
\textbf{Coloured-MNIST (ten-class)} - MNIST modified so that the digit colour is correlated with the class label. Hence, the input feature and spurious correlation are intertwined. We use all ten MNIST classes with ten unique colours, and the digit colour is perfectly correlated with the correct class label.

%
This experiment uses a convolutional neural network (CNN) comprised of two convolutional layers, with 32 and 64 filters, each followed by ReLUs, then $2\times 2$ max-pooling, a flattening layer, dropout layer, followed by two linear layers with 9216 and 128 neurons with dropout and ReLUs. The output is always a length-10 vector encoding the class label. 
We train from scratch our CNN on MNIST-SC, Coloured-MNIST, and standard MNIST, with and without our proposed ULE method. Then test on MNIST, which does not contain the spurious correlation. Our results are shown in \Cref{tab:proof_of_concept}.

When we train a neural network on MNIST-SC in the standard way, i.e., using ERM, we observe that the model focuses on the spurious correlation. It achieves perfect accuracy on both the training set and MNIST-SC testing set images with the spurious correlations present. However, the model's performance drops to 85\% when evaluated on MNIST testing set images, which do not contain spurious correlations. 
In contrast, when we train the same network using our proposed ULE method, it achieves an accuracy of 95\% whether or not the spurious correlation is present. This demonstrates that ULE has helped increase model robustness to spurious correlations.

When we train on ten-class Coloured-MNIST and test on ten-class MNIST, the performance of ULE is significantly better than ERM. This suggests that when the features are intertwined at the input, ULE can, to an extent, guide the teacher in ignoring spurious correlations.
Finally, we train and test on the standard MNIST dataset using ULE. In this case, no spurious correlations were present during training or testing. ULE's testing-set performance is only marginally below ERM, suggesting that ULE doesn't significantly harm performance, even in cases where the student has learned the correct features. 



\begin{table}[ht]
\centering
\caption{ULE vs baseline ERM. MNIST-SC, and Coloured-MNIST contain artificial spurious correlations. MNIST does not.\label{tab:proof_of_concept}}
\resizebox{\columnwidth}{!}{%
\begin{tabular}{@{}llcccc@{}}
\toprule
\multicolumn{1}{c}{Train Dataset} & \multicolumn{1}{c}{Test Dataset} & \multicolumn{2}{c}{Train Accuracy} & \multicolumn{2}{c}{Test Accuracy} \\ \midrule
\multicolumn{1}{c}{}   & \multicolumn{1}{c}{}   & ERM   & ULE (Ours)   & ERM   & ULE (Ours)  \\ \cmidrule(lr){3-4} \cmidrule(l){5-6}
MNIST                  & MNIST                  & 99\%    & 97\%    & 99\%    & 97\%    \\ \midrule
MNIST-SC               & MNIST-SC               & 100\%   & 98\%    & 100\%   & 95\%    \\
MNIST-SC               & MNIST                  & 100\%   & 98\%    & 85\%    & 95\%    \\ \midrule
ColoredMNIST           & ColoredMNIST           & 100\%   & 100\%   & 100\%   & 100\%   \\
ColoredMNIST           & MNIST                  & 100\%   & 100\%   & 21\%    & 40\%    \\ \bottomrule
\end{tabular}%
}
\end{table}

\begin{figure}[t!]
    \begin{subfigure}[t]{0.5\textwidth}
        \centering
        \includegraphics[height=1.5cm]{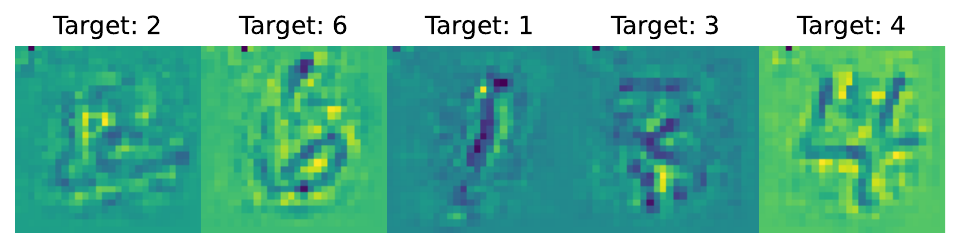}
        \caption{Raw gradients of ULE on MNIST-SC, showing clear focus on the digit.\label{fig:mnist_twin_grad}}        
    \end{subfigure}
    \hfill
    \begin{subfigure}[t]{0.5\textwidth}
        \centering
        \includegraphics[height=1.5cm]{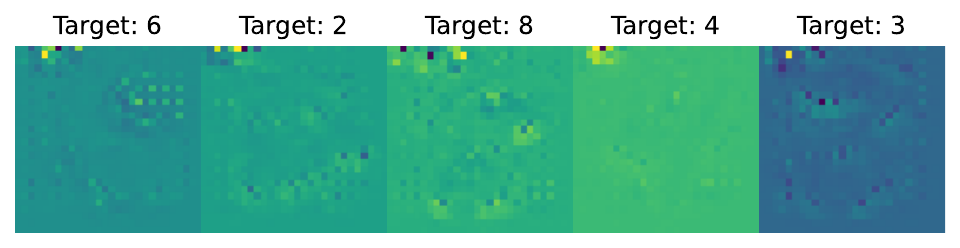}
        \caption{Raw gradients ERM on MNIST-SC, showing focus on hard-coded label.\label{fig:mnist_normal_grad}}
    \end{subfigure}
    \caption{Qualitative evaluation on MNIST-SC between gradients, $g_t(x)$, from our proposed ULE against an ERM baseline. Our proposed method, ULE, correctly focuses on the digits, whereas ERM focuses on the spurious correlation in the top-left corner.}
\end{figure}

To investigate further, we visualise $g_t(x)$, the gradient of the trained teacher network's output with respect to its input, for a random sample of images from the MNIST-SC testing set. \Cref{fig:mnist_normal_grad} shows that the ERM-trained model places significant attention on the upper left corner of the image, where the spurious correlation class labels were embedded. This shows the model learns to use the spurious correlation rather than focusing on the digits. In contrast, \Cref{fig:mnist_twin_grad}, shows our ULE model does not place emphasis on the top left corner but focuses on the digits instead. 
The results from \Cref{tab:proof_of_concept} and \Cref{fig:mnist_normal_grad,fig:mnist_twin_grad} show that our proposed ULE method can help train models that are robust to spurious correlations. With ULE the model does not rely on spurious correlations to achieve high accuracy even if they are clearly present in the data.

\subsection{Evaluating Spurious Correlation Robustness}
\label{subsec:evaluation}


In this section, we evaluate our ULE approach on several datasets with realistic spurious correlations and perform a comparison with state-of-the-art approaches. 

For each experiment, we tune the hyperparameters of our models as discussed above in \Cref{sec:experiments}. The learning rate, $L_2$ regularization and $\lambda$ hyperparameters were tuned independently for all datasets. 
The overall best-performing model on the validation set is saved and evaluated on the test set using $\overline{\text{Acc}}$ and $\text{WGA}$ 
(See \Cref{eq:average_test_accuracy,eq:worst_group_accuracy}). Training and evaluation of the models are repeated five times to compute the mean and standard deviation of the results. This ensures results are not affected by random initialization data shuffling. In this set of experiments, we use the same architecture for the student and teacher models. 

As discussed in \Cref{sec:efficient_computation}, we only fine-tune the final fully connected layer of the models. 
The gradients of the student model, $g_s$, 
are extracted from the output of the final block of convolutional layers for the ResNet and the output of the MLP Head for the ViT.
%
%
For all like-for-like comparison tables, we colour the \textbf{\textcolor{black}{1st}}, \textbf{\textcolor{blue}{2nd}} and \textbf{\textcolor{red}{3rd}} best results, and highlight results from other model architectures, which may not be direct comparable, \eg ViT-H-14 in \textbf{\textcolor{Gray}{Gray}}.

\subsubsection{Waterbirds}
\label{sec:waterbirds}
\Cref{tab:waterbirds_results} shows the results of our ULE method applied to three network architectures: ResNet-18, ResNet-50 and ViT-H-14. We compare our approach with state-of-art robustness methods on the Waterbirds dataset.  In a direct comparison, ULE-trained ResNet-50 equals the best result from the literature. Although not directly comparable, using the ViT-H-14 model, our method achieves a higher worst-group accuracy than all other current approaches, almost 2\% higher than the next best approach, even when compared against models that use group labels.

\begin{table}[ht]
\centering
\caption{Results of our method compared to recent approaches to robustness on the Waterbirds dataset, where $\dagger$ represents a paired model training and $\ast$ methods which make use of group labels in training or validation.\label{tab:waterbirds_results}}
\resizebox{\columnwidth}{!}{%
\begin{tabular}{@{}llcc@{}}
\toprule
Method & Model & Average Accuracy & \textbf{Worst-Group Accuracy} \\ \midrule
ERM~\cite{NIPS1991_ff4d5fbb} & ResNet-50 & 97.3\% & 60.0\% \\
EE~\cite{mehta2022you} & ViT-H-14 & 95.2\% & 90.1\% \\ \midrule
GroupDRO~\cite{sagawa2019distributionally}$\ast$ & ResNet-50 & 97.4\% & 86.0\% \\
LfF~\cite{nam2020learning}$\ast\dagger$ & ResNet-50 & 91.2\% & 78.0\% \\
EIIL~\cite{pmlr-v139-creager21a}$\ast$ & ResNet-50 & 96.9\% & \textbf{\textcolor{red}{78.7\%}} \\
JTT~\cite{pmlr-v139-liu21f}$\ast\dagger$ & ResNet-50 & 93.3\% & \textbf{\textcolor{blue}{86.7\%}} \\
SSA~\cite{nam2021spread}$\ast$ & ResNet-50 & 92.2\% & \textbf{\textcolor{black}{89.0\%}} \\
\midrule
ULE (Ours)$\dagger$ & ResNet-18 & 88.1\% & 87.7\% $\pm$ 0.01 \\
ULE (Ours)$\dagger$ & ResNet-50 & 89.6\% & \textbf{\textcolor{black}{89.0\% $\pm$ 0.02}} \\
\textcolor{Gray}{ULE (Ours)$\dagger$} & \textcolor{Gray}{ViT-H-14}  & \textcolor{Gray}{94.2\%} & \textcolor{Gray}{93.6\% $\pm$ 0.00} \\ \bottomrule
\end{tabular}%
}
\vspace{-0.5cm}
\end{table}


\subsubsection{XAI Analysis of Network trained on Waterbirds} 

\begin{figure}[t]
    \centering
    \includegraphics[width=0.9\columnwidth]{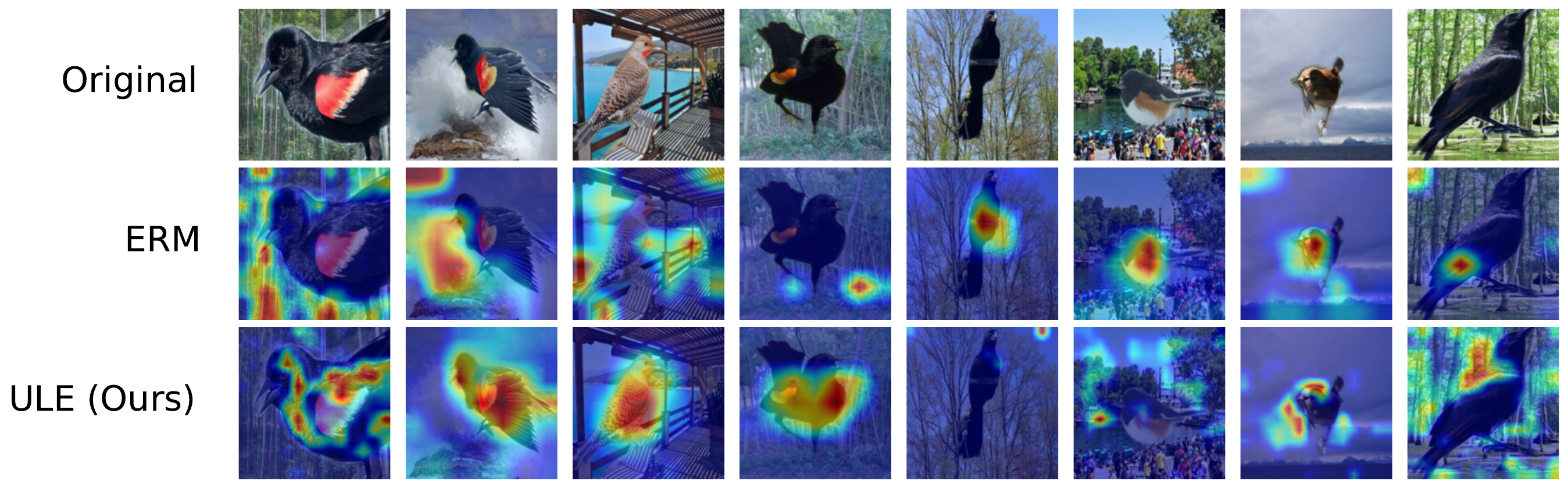}    
    \caption{Qualitative comparison of GradCAM heatmaps on Waterbirds from ULE vs ERM baseline. ULE tends to focus on the foreground and has learned to ignore background spurious correlations. Failure cases are shown in the four columns on the right.}
    \label{fig:gradcam_heatmaps}
    \vspace{-0.3cm}
\end{figure}

To compare the different behaviours of models trained with our proposed ULE vs an ERM baseline, we use the GradCAM~\cite{selvaraju2017grad} eXplainable AI method. We visualize the saliency maps of ULE and ERM, ResNet-50 models on a random selection of images from the Waterbirds test set. The results in~\Cref{fig:gradcam_heatmaps} show the ERM-trained model focuses on the background, i.e., the spurious correlations. In contrast, the ULE-trained model focuses on the subject, thus avoiding spurious correlations in the background. This is true across various images in \Cref{fig:gradcam_heatmaps}, demonstrating that our ULE method increases model robustness to spurious correlations, even in challenging realistic conditions. We also show some failure cases where the ERM model makes the wrong prediction. In these cases, it focuses on the background.


\subsubsection{Hyperparameter Sensitivity}

\begin{figure}[t]
    \centering
    \includegraphics[width=\columnwidth]{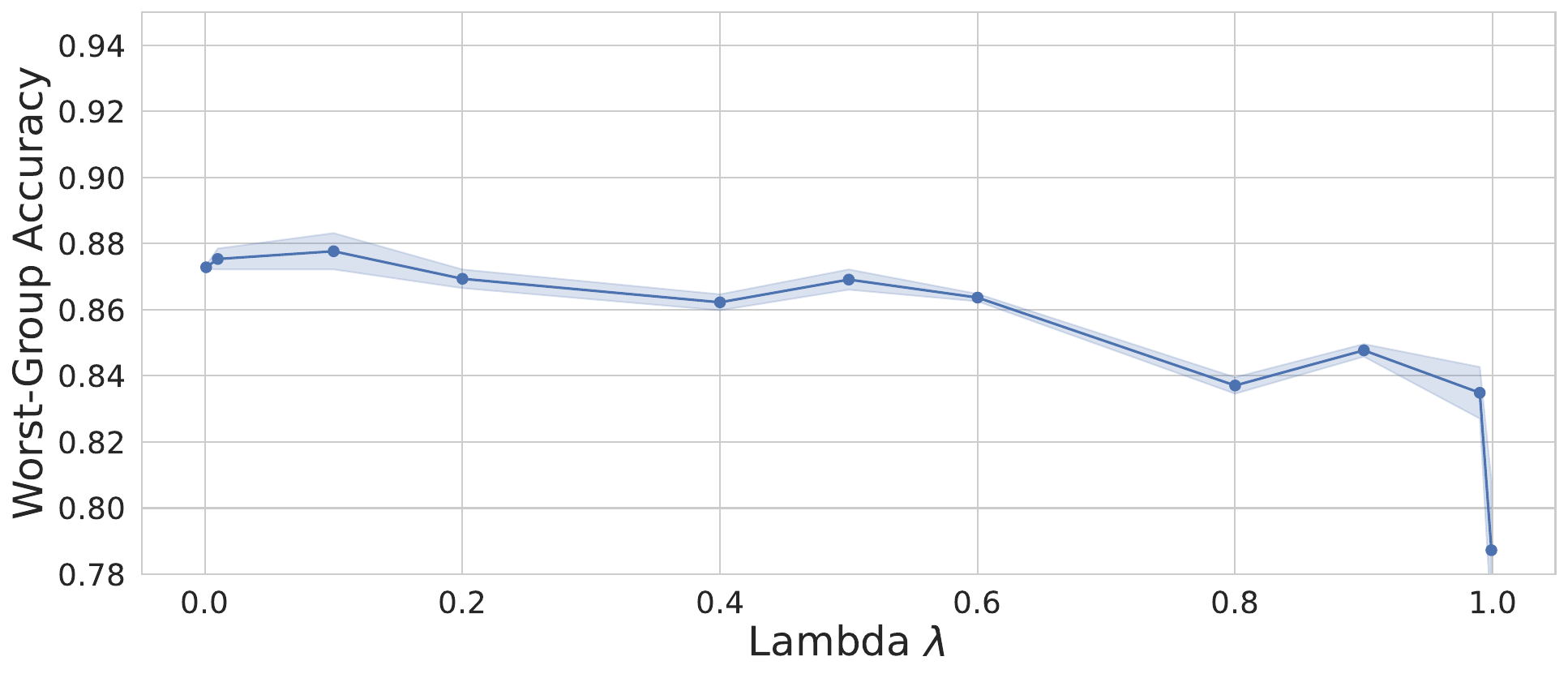}
    \caption{Sensitivity of ULE to changes in $\lambda$, tested on Waterbirds.}
    \label{fig:lambdaWGAWaterbirds}
    \vspace{-0.3cm}
\end{figure}

The loss function, Eq.~\ref{eq:overall_loss_function}, is critical to functionality ULE. Therefore, we measure the sensitivity of ULE to changes in the value of the $\lambda$ hyperparameter, which controls the balance between the classification and gradient loss terms of the teacher network. For each value of $\lambda$, we trained a ResNet-50 model with ULE on Waterbirds 3 times and averaged the WGA. The results in Fig.~\ref{fig:lambdaWGAWaterbirds} indicate that ULE performs well over a wide range of $\lambda$ values.

Using the same procedure, we also varied the second component of the loss function in Eq.~\ref{eq:overall_loss_function} to use L1 loss rather than MSE loss. This resulted in a WGA of 87.3 and an average accuracy of 89.0. These values are marginally below those in Table~\ref{tab:waterbirds_results}, indicating that the choice of MSE or L1 loss is not critical to the functionality of ULE.

\subsubsection{Waterbirds -- Trained from Scratch}

In common with the majority of other spurious correlation robustness methods~\cite{joshi2023towards}, ULE typically acts on the final layer representations from a pre-trained model. However, ULE can also be used when training a model from scratch (Also see Section~\ref{subsec:proof_of_concept}). We train a ResNet-50 model from scratch on Waterbirds using ULE, where the saliency, $\partial t(x) / \partial x$, is taken with respect to the input image, and all network layers are allowed to train. 

In Table~\ref{tab:waterbirds_from_scratch}, we contrast ULE with the comparable from-scratch results reported in~\cite{joshi2023towards}. In common with other methods, ULE's from-scratch performance is worse than when using a pre-trained model (See Section~\ref{sec:waterbirds}). However, ULE outperforms ERM, and is comparable with the best other methods in terms of average accuracy and WGA. 

\begin{table}[ht]
\centering
\caption{Comparison of ULE trained from scratch on Waterbirds with the results from~\cite{joshi2023towards} under the same experimental conditions. \label{tab:waterbirds_from_scratch}}
\vspace{-5mm}
\resizebox{\columnwidth}{!}{%
\begin{tabular}{@{}llcc@{}}
\toprule
Method & Model & Average Accuracy & \textbf{Worst-Group Accuracy} \\ \midrule
ERM~\cite{NIPS1991_ff4d5fbb}      & ResNet-50 			& 82.2\% $\pm$ 6.4 	& 23.5\% $\pm$ 5.0 \\ \midrule
CB~\cite{joshi2023towards}       & ResNet-50 			& 77.4\% $\pm$ 7.8 	& 28.4\% $\pm$ 4.7 \\
EIIL~\cite{pmlr-v139-creager21a}     & ResNet-50 		& 51.1\% $\pm$ 3.2 	& 45.5\% $\pm$ 3.2 \\
JTT~\cite{pmlr-v139-liu21f}      & ResNet-50 			& 85.7\% $\pm$ 0.4 	& 15.7\% $\pm$ 1.6 \\
Spare~\cite{yang2024identifying}    & ResNet-50 		& 59.5\% $\pm$ 4.9 	& 50.6\% $\pm$ 1.3 \\
SSA~\cite{nam2021spread}      & ResNet-50 				& 62.1\% $\pm$ 8.2 	& 47.8\% $\pm$ 1.7 \\
DFR~\cite{kirichenko2022last}      & ResNet-50 			& 63.8\% $\pm$ 1.1 	& \textbf{\textcolor{blue}{50.8\% $\pm$ 1.8}} \\
Dispel~\cite{xue2023few}   & ResNet-50 					& 65.2\% $\pm$ 1.6 	& \textbf{\textcolor{black}{51.9\% $\pm$ 1.2}} \\
GB~\cite{joshi2023towards}       & ResNet-50 			& 64.3\% $\pm$ 2.1 	& 50.6 \% $\pm$ 1.1 \\ 
GroupDRO~\cite{sagawa2019distributionally} & ResNet-50 	& 65.1\% $\pm$ 1.2 	& \textbf{\textcolor{red}{50.7\% $\pm$ 0.7}} \\ \midrule
ULE (Ours)     & ResNet-50 								& 62.8\% $\pm$ 1.4  & \textbf{\textcolor{red}{50.7\% $\pm$ 0.9}} \\ \bottomrule
\end{tabular}%
}
\vspace{-0.1cm}
\end{table}

\subsubsection{CelebA}
\Cref{tab:celeba_results} compares ULE with state-of-art approaches to spurious correlation robustness on the CelebA dataset. Our ULE method is in the top-3 best methods in terms of worst-group accuracy. Only SSA~\cite{nam2021spread} and GroupDRO~\cite{sagawa2019distributionally} achieve higher worst-group accuracy. It is important to note that our method does not use group labels in training or validation, while those other techniques do, meaning it is easier to apply our approach in new situations.


\begin{table}[ht]
\centering
\caption{ULE vs recent methods on the CelebA dataset. Paired model training $\dagger$. Group labels in training or validation $\ast$.\label{tab:celeba_results}}
\vspace{-2mm}
\resizebox{\columnwidth}{!}{%
\begin{tabular}{@{}llcc@{}}
\toprule
Method & Model & Average Accuracy & \textbf{Worst-Group Accuracy} \\ \midrule
ERM~\cite{NIPS1991_ff4d5fbb} & ResNet-50 & 94.8\% & 41.1\% \\ \midrule
GroupDRO~\cite{sagawa2019distributionally}$\ast$ & ResNet-50 & 91.8\% & \textbf{\textcolor{blue}{88.3\%}} \\
LfF~\cite{nam2020learning}$\ast\dagger$ & ResNet-50 & 86.0\% & 70.6\% \\
EIIL~\cite{pmlr-v139-creager21a}$\ast$ & ResNet-50 & 91.9\% & 83.3\% \\
JTT~\cite{pmlr-v139-liu21f}$\ast\dagger$ & ResNet-50 & 88.0\% & 81.1\% \\
SSA~\cite{nam2021spread}$\ast$ & ResNet-50 & 92.8\% & \textbf{\textcolor{black}{89.8\%}} \\ \midrule
ULE (Ours)$\dagger$ & ResNet-18 & 85.7\% & 84.6\% $\pm$ 0.02 \\
ULE (Ours)$\dagger$ & ResNet-50 & 87.6\% & \textbf{\textcolor{red}{85.3\% $\pm$ 0.01}} \\
\textcolor{Gray}{ULE (Ours)$\dagger$} & \textcolor{Gray}{ViT-H-14} & \textcolor{Gray}{88.3\%} & \textcolor{Gray}{87.1\%} \textcolor{Gray}{$\pm$ 0.00} \\ \bottomrule
\end{tabular}%
}
\vspace{-0.5cm}
\end{table}

\subsubsection{Spawrious} Spawrious is more complex than Waterbirds or CelebA. It includes one-to-one and many-to-many spurious correlations. Following the procedure in Lynch et al.~\cite{lynch2023spawrious}, 
in \Cref{tab:spawrious_m2m_o2o_results} we evaluate using average accuracy to allow for direct comparison against the results reported in~\cite{lynch2023spawrious}. Additionally, in Table~\ref{tab:spawrious_ule}, we report WGA for consistency and to allow future comparison with our method. We are unaware of any WGA results in the literature for this dataset, so we make no claims about ULE's performance compared to other methods. 

\textbf{Spawrious: One-to-One} 
In \Cref{tab:spawrious_m2m_o2o_results}, we compare ULE against the literature on Spawrious: One-to-One. 
ULE with the ResNet-50 model achieves the highest average accuracy on the easy setting and is competitive on other difficulties. 

\textbf{Spawrious: Many-to-Many}
Many-to-many spurious correlations happen when the spurious correlations hold over disjoint groups of spurious attributes and classes. For instance, each class from the group \{Bulldog, Dachshund\} is observed with each background from the group \{Desert, Jungle\} in equal proportion in the training set~\cite{lynch2023spawrious}. These more complex scenarios require the model to learn to ignore spurious correlations for more than one class and group combination. Robustness to many-to-many spurious correlations is important because they can occur in real settings.

\begin{table}[ht!]
\centering
\caption{ULE ResNet-50 pre-trained on ImageNet1K\_V2 compared to recent approaches on Spawrious: One-to-One and Many-to-Many using ResNet-50. Results reproduced from~\cite{lynch2023spawrious}.\label{tab:spawrious_m2m_o2o_results}}
\vspace{-2mm}
\resizebox{\columnwidth}{!}{%
\begin{tabular}{@{}lccccccc@{}}
\toprule
\multicolumn{1}{c}{} & \multicolumn{3}{c}{One-to-One} & \multicolumn{3}{c}{Many-to-Many} \\ \cmidrule(lr){2-4} \cmidrule(lr){5-7} \cmidrule(l){8-8}
Method & Easy & Medium & Hard & Easy & Medium & Hard & Average \\ \midrule
ERM~\cite{NIPS1991_ff4d5fbb} & 77.49\% & 76.60\% & 71.32\% & 83.80\% & 53.05\% & 58.70\% & 70.16\% \\
GroupDRO~\cite{sagawa2019distributionally} & 80.58\% & 75.96\% & 76.99\% & 79.96\% & 61.01\% & 60.86\% & 72.56\% \\
IRM~\cite{arjovsky2019invariant} & 75.45\% & 76.39\% & 74.90\% & 76.15\% & 67.82\% & 60.93\% & 71.94\% \\
Coral~\cite{dcoral} & \textbf{\textcolor{red}{89.66\%}} & \textbf{\textcolor{red}{81.05\%}} & 79.65\% & 81.26\% & 65.18\% & 67.97\% & 77.46\% \\
CausIRL~\cite{chevalley2022invariant} & 89.32\% & 78.64\% & \textbf{\textcolor{red}{80.40\%}} & 86.44\% & 66.11\% & \textbf{\textcolor{red}{71.36\%}} & 77.20\% \\
MMD-AAE~\cite{Li_2018_CVPR} & 78.81\% & 75.33\% & 72.66\% & 78.91\% & 64.21\% & 66.86\% & 70.20\% \\
Fish~\cite{shi2021gradient} & 77.51\% & 77.72\% & 74.73\% & 81.60\% & 63.03\% & 58.94\% & 72.26\% \\
VREx~\cite{krueger2021out} & 84.69\% & 77.56\% & 75.41\% & 81.22\% & 54.28\% & 59.21\% & 72.06\% \\
W2D~\cite{huang2022two} & 81.94\% & 76.74\% & 76.84\% & 80.80\% & 62.82\% & 61.89\% & 73.50\% \\
JTT~\cite{pmlr-v139-liu21f} & \textbf{\textcolor{blue}{90.24\%}} & \textbf{\textcolor{black}{87.28\%}} & \textbf{\textcolor{black}{87.41\%}} & 79.23\% & 60.56\% & 57.58\% & 77.05\% \\
Mixup-RS~\cite{xu2020adversarial} & 88.48\% & \textbf{\textcolor{blue}{82.75\%}} & 75.75\% & \textbf{\textcolor{blue}{89.61\%}} & \textbf{\textcolor{blue}{77.23\%}} & 71.21\% & \textbf{\textcolor{blue}{80.84\%}} \\
Mixup-LISA~\cite{yao2022improving} & 88.64\% & 80.83\% & 72.54\% & \textbf{\textcolor{red}{87.24\%}} & \textbf{\textcolor{red}{71.78\%}} & \textbf{\textcolor{blue}{72.97\%}} & \textbf{\textcolor{red}{79.00\%}} \\ \midrule
ULE (Ours) & \textbf{\textcolor{black}{92.00\%}} & 75.39\% & 76.76\% & \textbf{\textcolor{black}{90.61\%}} & \textbf{\textcolor{black}{82.43\%}} & \textbf{\textcolor{black}{80.37\%}} & \textbf{\textcolor{black}{82.00\%}} \\ \bottomrule
\end{tabular}%
}
\end{table}

\begin{table}[ht!]
\centering
\caption{WGA of ULE on Spawrious with different models. ERM results are also provided as reference.\label{tab:spawrious_ule}}
\vspace{-2.5mm}
\resizebox{\columnwidth}{!}{%
\begin{tabular}{@{}llcccccccc@{}}
\toprule
\multicolumn{1}{c}{} & \multicolumn{1}{c}{} & \multicolumn{4}{c}{One-to-One} & \multicolumn{4}{c}{Many-to-Many} \\ \cmidrule(lr){3-6} \cmidrule(l){7-10}
Method & Model & Easy & Medium & Hard & Average & Easy & Medium & Hard & Average \\ \midrule
ERM & ResNet-18 & 66.20\% & 45.80\% & 39.50\% & 50.50\% & 73.80\% & 58.90\% & 54.90\% & 62.50\% \\ \midrule
ULE & ResNet-18 & 79.43\% & 61.38\% & 56.31\% & 65.71\% & 82.81\% & 65.48\% & 61.24\% & 69.84\% \\
ULE & ResNet-50 & 87.03\% & 54.45\% & 66.37\% & 69.28\% & 87.67\% & 75.95\% & 73.60\% & 79.07\% \\
ULE & ViT-H-14 & 90.61\% & 89.27\% & 85.60\% & 88.49\% & 94.60\% & 88.94\% & 89.99\% & 91.18\% \\ \bottomrule
\end{tabular}%
}
\end{table}


\Cref{tab:spawrious_m2m_o2o_results} shows a comparison between ULE and state-of-the-art methods on the Spawrious: Many-to-Many benchmark. 
ULE with ResNet-50 achieves the highest average accuracy across all difficulty settings.


\subsubsection{UrbanCars}

UrbanCars~\cite{li2023whac} is a challenging dataset that includes multiple types of spurious correlations. Both the image background and a co-occurring object are correlated with the true class. We follow the protocol of Li et al.~\cite{li2023whac}, which allows for direct comparison with results from the literature. 

In \Cref{tab:urbancars_results}, we see that ULE performs in the top three of recently published methods in terms of WGA. We also see that ULE has the highest average accuracy of all methods.

\begin{table}[ht]
\centering
\caption{ULE compared to recent approaches on the UrbanCars dataset, where $\dagger$ represents a paired model training and $\ast$ methods which make use of group labels in training or validation.\label{tab:urbancars_results}}
\resizebox{\columnwidth}{!}{%
\begin{tabular}{@{}llcc@{}}
\toprule
Method & Model & Average Accuracy & \textbf{Worst-Group Accuracy} \\ \midrule
ERM~\cite{NIPS1991_ff4d5fbb} & ResNet-50 & 97.6\% & 28.4\% \\ \midrule
GroupDRO~\cite{sagawa2019distributionally}$\ast$ & ResNet-50 & 91.6\% & \textbf{\textcolor{blue}{75.2\%}} \\
LfF~\cite{nam2020learning}$\ast\dagger$ & ResNet-50 & 97.2\% & 34.0\% \\
EIIL~\cite{pmlr-v139-creager21a}$\ast$ & ResNet-50 & 95.5\% & 50.6\% \\
JTT~\cite{pmlr-v139-liu21f}$\ast\dagger$ & ResNet-50 & 95.9\% & 55.8\% \\
SPARE~\cite{yang2024identifying}$\ast$ & ResNet-50 & 96.6\% & \textbf{\textcolor{black}{76.9\%}} \\
\midrule
ULE (Ours)$\dagger$ & ResNet-50 & 98.2\% & \textbf{\textcolor{red}{71.6\%}} \\ \bottomrule
\end{tabular}%
}
\vspace{-0.3cm}
\end{table}

\subsection{Mixed Models}
\label{subsec:mixed_models}


We now investigate ULE's performance when different model architectures are used for teacher and student. We compare performance when models are paired with themselves versus paired with other models.
%
%
In common with~\Cref{sec:experiments}, we perform hyperparameter tuning to select the best hyperparameters for each model combination. All experiments were performed on the Waterbirds dataset.


\begin{table}[ht]
\centering
\caption{Waterbirds WGA for combinations of student \& teacher.\label{tab:mixed_model_results}}
\vspace{-1mm}
\resizebox{\columnwidth}{!}{
\begin{tabular}{@{}llll@{}}
\toprule
\backslashbox[0pt][l]{Teacher}{Student} & ResNet-18 & ResNet-50 & ViT-H-14 \\ \midrule
ResNet-18 & 87.7\% $\pm$ 0.01      & 87.4\% $\pm$ 0.01      & 87.6\% $\pm$ 0.01 \\
ResNet-50 & 88.7\% $\pm$ 0.02      & 89.0\% $\pm$ 0.02      & 87.5\% $\pm$ 0.01 \\
ViT-H-14  & 91.7\% $\pm$ 0.00      & 91.9\% $\pm$ 0.01      & 93.6\% $\pm$ 0.00 \\ \bottomrule
\end{tabular}%
}
\end{table}

\Cref{tab:mixed_model_results} shows worst-group test accuracies for different teacher and student model architecture combinations. The results show a correlation between the total complexity of the teacher and student models and worst-group accuracy. As the total complexity increases, the worst-group accuracy increases consistently, with the most complex combination, ViT-H-14 paired with itself, achieving the highest worst-group accuracy. When using mixed models, we also see that having a simpler student model and a more complex teacher model leads to better performance. Indeed, since the student aims to highlight mistakes to the teacher, its complexity or architecture seems to be of little importance. Our intuition is that the teacher model requires more capacity to be capable of solving the task in a different way than the student model, thus learning to ignore the spurious correlations. The complexity of the teacher seems to be the main factor in determining performance, with the ViT-H-14 teacher clearly outperforming any other choice.




\section{Conclusion}
\label{sec:conclusion}

We propose UnLearning from Experience (ULE), a new twist on student-teacher approaches that reverses the usual roles of student and teacher. The teacher observes the gradients of the student model, and ensures its gradients are the opposite, hence ``unlearning'' the student's mistakes to increase its robustness to spurious correlations.  
We demonstrate the effectiveness of ULE on the, Waterbirds, CelebA, and Spawrious datasets. ULE achieves state-of-the-art results on Waterbirds and CelebA, and is competitive on Spawrious. 
ULE does not require prior knowledge of the spurious correlations and is not affected when spurious correlations are not present. It does not require group labels, unlike some other approaches (\Cref{sec:literature_review}). ULE is simple to implement and can be applied to many model architectures. These factors enhance its real-world applicability.

{\small
\bibliographystyle{ieee_fullname}
\bibliography{egbib}

\begin{thebibliography}{10}\itemsep=-1pt

\bibitem{arjovsky2019invariant}
Martin Arjovsky, L{\'e}on Bottou, Ishaan Gulrajani, and David Lopez-Paz.
\newblock Invariant risk minimization.
\newblock {\em arXiv preprint arXiv:1907.02893}, 2019.

\bibitem{chevalley2022invariant}
Mathieu Chevalley, Charlotte Bunne, Andreas Krause, and Stefan Bauer.
\newblock Invariant causal mechanisms through distribution matching.
\newblock {\em arXiv preprint arXiv:2206.11646}, 2022.

\bibitem{pmlr-v139-creager21a}
Elliot Creager, Joern-Henrik Jacobsen, and Richard Zemel.
\newblock Environment inference for invariant learning.
\newblock In Marina Meila and Tong Zhang, editors, {\em Proceedings of the 38th International Conference on Machine Learning}, volume 139 of {\em Proceedings of Machine Learning Research}, pages 2189--2200. PMLR, 18--24 Jul 2021.

\bibitem{darbinyan2023identifying}
Rafayel Darbinyan, Hrayr Harutyunyan, Aram~H Markosyan, and Hrant Khachatrian.
\newblock Identifying and disentangling spurious features in pretrained image representations.
\newblock {\em arXiv preprint arXiv:2306.12673}, 2023.

\bibitem{deng2012mnist}
Li Deng.
\newblock The mnist database of handwritten digit images for machine learning research.
\newblock {\em IEEE Signal Processing Magazine}, 29(6):141--142, 2012.

\bibitem{dosovitskiy2020image}
Alexey Dosovitskiy, Lucas Beyer, Alexander Kolesnikov, Dirk Weissenborn, Xiaohua Zhai, Thomas Unterthiner, Mostafa Dehghani, Matthias Minderer, Georg Heigold, Sylvain Gelly, et~al.
\newblock An image is worth 16x16 words: Transformers for image recognition at scale.
\newblock In {\em International Conference on Learning Representations}, 2020.

\bibitem{he2016deep}
Kaiming He, Xiangyu Zhang, Shaoqing Ren, and Jian Sun.
\newblock Deep residual learning for image recognition.
\newblock In {\em Proceedings of the IEEE conference on computer vision and pattern recognition}, pages 770--778, 2016.

\bibitem{huang2022two}
Zeyi Huang, Haohan Wang, Dong Huang, Yong~Jae Lee, and Eric~P Xing.
\newblock The two dimensions of worst-case training and their integrated effect for out-of-domain generalization.
\newblock In {\em Proceedings of the IEEE/CVF Conference on Computer Vision and Pattern Recognition}, pages 9631--9641, 2022.

\bibitem{izmailov2022feature}
Pavel Izmailov, Polina Kirichenko, Nate Gruver, and Andrew~G Wilson.
\newblock On feature learning in the presence of spurious correlations.
\newblock {\em NeurIPS}, 35:38516--38532, 2022.

\bibitem{joshi2023towards}
Siddharth Joshi, Yu Yang, Yihao Xue, Wenhan Yang, and Baharan Mirzasoleiman.
\newblock Towards mitigating spurious correlations in the wild: A benchmark \& a more realistic dataset.
\newblock {\em arXiv preprint arXiv:2306.11957}, 2023.

\bibitem{kim2022efficient}
Junyaup Kim and Simon~S Woo.
\newblock Efficient two-stage model retraining for machine unlearning.
\newblock In {\em Proceedings of the IEEE/CVF Conference on Computer Vision and Pattern Recognition}, pages 4361--4369, 2022.

\bibitem{kirichenko2022last}
Polina Kirichenko, Pavel Izmailov, and Andrew~Gordon Wilson.
\newblock Last layer re-training is sufficient for robustness to spurious correlations.
\newblock In {\em The Eleventh International Conference on Learning Representations}, 2022.

\bibitem{krueger2021out}
David Krueger, Ethan Caballero, Joern-Henrik Jacobsen, Amy Zhang, Jonathan Binas, Dinghuai Zhang, Remi Le~Priol, and Aaron Courville.
\newblock Out-of-distribution generalization via risk extrapolation (rex).
\newblock In {\em International conference on machine learning}, pages 5815--5826. PMLR, 2021.

\bibitem{lee2022diversify}
Yoonho Lee, Huaxiu Yao, and Chelsea Finn.
\newblock Diversify and disambiguate: Learning from underspecified data.
\newblock In {\em ICML 2022: Workshop on Spurious Correlations, Invariance and Stability}, 2022.

\bibitem{Li_2018_CVPR}
Haoliang Li, Sinno~Jialin Pan, Shiqi Wang, and Alex~C. Kot.
\newblock Domain generalization with adversarial feature learning.
\newblock In {\em Proceedings of the IEEE Conference on Computer Vision and Pattern Recognition (CVPR)}, June 2018.

\bibitem{li2023whac}
Zhiheng Li, Ivan Evtimov, Albert Gordo, Caner Hazirbas, Tal Hassner, Cristian~Canton Ferrer, Chenliang Xu, and Mark Ibrahim.
\newblock A whac-a-mole dilemma: Shortcuts come in multiples where mitigating one amplifies others.
\newblock In {\em Proceedings of the IEEE/CVF Conference on Computer Vision and Pattern Recognition}, pages 20071--20082, 2023.

\bibitem{lin2022zin}
Yong Lin, Shengyu Zhu, Lu Tan, and Peng Cui.
\newblock Zin: When and how to learn invariance without environment partition?
\newblock {\em Advances in Neural Information Processing Systems}, 35:24529--24542, 2022.

\bibitem{pmlr-v139-liu21f}
Evan~Z Liu, Behzad Haghgoo, Annie~S Chen, Aditi Raghunathan, Pang~Wei Koh, Shiori Sagawa, Percy Liang, and Chelsea Finn.
\newblock Just train twice: Improving group robustness without training group information.
\newblock In Marina Meila and Tong Zhang, editors, {\em Proceedings of the 38th International Conference on Machine Learning}, volume 139 of {\em Proceedings of Machine Learning Research}, pages 6781--6792. PMLR, 18--24 Jul 2021.

\bibitem{liu2015faceattributes}
Ziwei Liu, Ping Luo, Xiaogang Wang, and Xiaoou Tang.
\newblock Deep learning face attributes in the wild.
\newblock In {\em Proceedings of International Conference on Computer Vision (ICCV)}, 2015.

\bibitem{lynch2023spawrious}
Aengus Lynch, Gbètondji J-S Dovonon, Jean Kaddour, and Ricardo Silva.
\newblock Spawrious: A benchmark for fine control of spurious correlation biases, 2023.

\bibitem{mehta2022you}
Raghav Mehta, V{\'\i}tor Albiero, Li Chen, Ivan Evtimov, Tamar Glaser, Zhiheng Li, and Tal Hassner.
\newblock You only need a good embeddings extractor to fix spurious correlations.
\newblock {\em arXiv preprint arXiv:2212.06254}, 2022.

\bibitem{nam2020learning}
Junhyun Nam, Hyuntak Cha, Sungsoo Ahn, Jaeho Lee, and Jinwoo Shin.
\newblock Learning from failure: training debiased classifier from biased classifier.
\newblock In {\em Proceedings of the 34th International Conference on Neural Information Processing Systems}, pages 20673--20684, 2020.

\bibitem{nam2021spread}
Junhyun Nam, Jaehyung Kim, Jaeho Lee, and Jinwoo Shin.
\newblock Spread spurious attribute: Improving worst-group accuracy with spurious attribute estimation.
\newblock In {\em International Conference on Learning Representations}, 2021.

\bibitem{NEURIPS2022_baaa7b5b}
Hongjing Niu, Hanting Li, Feng Zhao, and Bin Li.
\newblock Roadblocks for temporarily disabling shortcuts and learning new knowledge.
\newblock In S. Koyejo, S. Mohamed, A. Agarwal, D. Belgrave, K. Cho, and A. Oh, editors, {\em Advances in Neural Information Processing Systems}, volume~35, pages 29064--29075. Curran Associates, Inc., 2022.

\bibitem{pagliardini2022agree}
Matteo Pagliardini, Martin Jaggi, Fran{\c{c}}ois Fleuret, and Sai~Praneeth Karimireddy.
\newblock Agree to disagree: Diversity through disagreement for better transferability.
\newblock In {\em ICLR}, 2022.

\bibitem{qiu2023afr}
Shikai Qiu, Andres Potapczynski, Pavel Izmailov, and Andrew~Gordon Wilson.
\newblock {Simple and Fast Group Robustness by Automatic Feature Reweighting}.
\newblock {\em International Conference on Machine Learning (ICML)}, 2023.

\bibitem{sagawa2019distributionally}
Shiori Sagawa, Pang~Wei Koh, Tatsunori~B Hashimoto, and Percy Liang.
\newblock Distributionally robust neural networks.
\newblock In {\em International Conference on Learning Representations}, 2019.

\bibitem{selvaraju2017grad}
Ramprasaath~R Selvaraju, Michael Cogswell, Abhishek Das, Ramakrishna Vedantam, Devi Parikh, and Dhruv Batra.
\newblock Grad-cam: Visual explanations from deep networks via gradient-based localization.
\newblock In {\em Proceedings of the IEEE International Conference on Computer Vision}, pages 618--626, 2017.

\bibitem{shah2020pitfalls}
Harshay Shah, Kaustav Tamuly, Aditi Raghunathan, Prateek Jain, and Praneeth Netrapalli.
\newblock The pitfalls of simplicity bias in neural networks.
\newblock {\em Advances in Neural Information Processing Systems}, 33:9573--9585, 2020.

\bibitem{sharif2014cnn}
Ali Sharif~Razavian, Hossein Azizpour, Josephine Sullivan, and Stefan Carlsson.
\newblock Cnn features off-the-shelf: an astounding baseline for recognition.
\newblock In {\em Proceedings of the IEEE conference on computer vision and pattern recognition workshops}, pages 806--813, 2014.

\bibitem{shi2021gradient}
Yuge Shi, Jeffrey Seely, Philip~HS Torr, N Siddharth, Awni Hannun, Nicolas Usunier, and Gabriel Synnaeve.
\newblock Gradient matching for domain generalization.
\newblock {\em arXiv preprint arXiv:2104.09937}, 2021.

\bibitem{singh2022revisiting}
Mannat Singh, Laura Gustafson, Aaron Adcock, Vinicius de~Freitas Reis, Bugra Gedik, Raj~Prateek Kosaraju, Dhruv Mahajan, Ross Girshick, Piotr Doll{\'a}r, and Laurens van~der Maaten.
\newblock {Revisiting Weakly Supervised Pre-Training of Visual Perception Models}.
\newblock In {\em CVPR}, 2022.

\bibitem{srivastava2023mitigating}
Mashrin Srivastava.
\newblock Mitigating spurious correlations in machine learning models: Techniques and applications.
\newblock 2023.

\bibitem{coral}
Baochen Sun, Jiashi Feng, and Kate Saenko.
\newblock Return of frustratingly easy domain adaptation.
\newblock In {\em AAAI}, 2016.

\bibitem{dcoral}
Baochen Sun and Kate Saenko.
\newblock Deep coral: Correlation alignment for deep domain adaptation.
\newblock In {\em ECCV 2016 Workshops}, 2016.

\bibitem{pmlr-v202-tiwari23a}
Rishabh Tiwari and Pradeep Shenoy.
\newblock Overcoming simplicity bias in deep networks using a feature sieve.
\newblock In Andreas Krause, Emma Brunskill, Kyunghyun Cho, Barbara Engelhardt, Sivan Sabato, and Jonathan Scarlett, editors, {\em Proceedings of the 40th International Conference on Machine Learning}, volume 202 of {\em Proceedings of Machine Learning Research}, pages 34330--34343. PMLR, 23--29 Jul 2023.

\bibitem{NIPS1991_ff4d5fbb}
V. Vapnik.
\newblock Principles of risk minimization for learning theory.
\newblock In J. Moody, S. Hanson, and R.P. Lippmann, editors, {\em Advances in Neural Information Processing Systems}, volume~4. Morgan-Kaufmann, 1991.

\bibitem{wah2011caltech}
Catherine Wah, Steve Branson, Peter Welinder, Pietro Perona, and Serge Belongie.
\newblock The caltech-ucsd birds-200-2011 dataset.
\newblock In {\em Technical Report CNS-TR-2011-001, California Institute of Technology}, 2011.

\bibitem{xie2020self}
Qizhe Xie, Minh-Thang Luong, Eduard Hovy, and Quoc~V Le.
\newblock Self-training with noisy student improves imagenet classification.
\newblock In {\em Proceedings of the IEEE/CVF conference on computer vision and pattern recognition}, pages 10687--10698, 2020.

\bibitem{xu2020adversarial}
Minghao Xu, Jian Zhang, Bingbing Ni, Teng Li, Chengjie Wang, Qi Tian, and Wenjun Zhang.
\newblock Adversarial domain adaptation with domain mixup.
\newblock In {\em Proceedings of the AAAI conference on artificial intelligence}, volume~34, pages 6502--6509, 2020.

\bibitem{xue2023few}
Yihao Xue, Ali Payani, Yu Yang, and Baharan Mirzasoleiman.
\newblock Few-shot adaption to distribution shifts by mixing source and target embeddings.
\newblock {\em arXiv preprint arXiv:2305.14521}, 2023.

\bibitem{yang2024identifying}
Yu Yang, Eric Gan, Gintare~Karolina Dziugaite, and Baharan Mirzasoleiman.
\newblock Identifying spurious biases early in training through the lens of simplicity bias.
\newblock In {\em International Conference on Artificial Intelligence and Statistics}, pages 2953--2961. PMLR, 2024.

\bibitem{yao2022improving}
Huaxiu Yao, Yu Wang, Sai Li, Linjun Zhang, Weixin Liang, James Zou, and Chelsea Finn.
\newblock Improving out-of-distribution robustness via selective augmentation.
\newblock In {\em International Conference on Machine Learning}, pages 25407--25437. PMLR, 2022.

\bibitem{zhang2022correct}
Michael Zhang, Nimit~S Sohoni, Hongyang~R Zhang, Chelsea Finn, and Christopher Re.
\newblock Correct-n-contrast: a contrastive approach for improving robustness to spurious correlations.
\newblock In {\em International Conference on Machine Learning}, pages 26484--26516. PMLR, 2022.

\bibitem{zhang2018generalized}
Zhilu Zhang and Mert~R Sabuncu.
\newblock Generalized cross-entropy loss for training deep neural networks with noisy labels.
\newblock In {\em Proceedings of the 32nd International Conference on Neural Information Processing Systems}, pages 8792--8802, 2018.

\bibitem{zhou2018places}
Bolei Zhou, Agata Lapedriza, Aditya Khosla, Aude Oliva, and Antonio Torralba.
\newblock Places: A 10 million image database for scene recognition.
\newblock {\em IEEE Transactions on Pattern Analysis \& Machine Intelligence}, 40(06):1452--1464, 2018.

\end{thebibliography}
}

\end{document}